\documentclass[10pt,twocolumn,letterpaper]{article}

\usepackage{wacv}              % To produce the CAMERA-READY version
\usepackage{graphicx}
\usepackage{amsmath}
\usepackage{amssymb}
\usepackage{booktabs}

\usepackage[pagebackref,breaklinks,colorlinks]{hyperref}

\usepackage[capitalize]{cleveref}
\crefname{section}{Sec.}{Secs.}
\Crefname{section}{Section}{Sections}
\Crefname{table}{Table}{Tables}
\crefname{table}{Tab.}{Tabs.}

\usepackage{comment}
\usepackage{subcaption}
\usepackage{placeins}
\usepackage{bm}
\usepackage{listings}
\usepackage{xcolor}
\usepackage{multirow}

\definecolor{codegreen}{rgb}{0,0,1}
\definecolor{codegray}{rgb}{0.95,0.5,0.5}
\definecolor{codepurple}{rgb}{0.58,0,0.82}
\definecolor{backcolour}{rgb}{1, 1, 1}
\usepackage{times}
\usepackage{epsfig}
\usepackage{graphicx}
\usepackage{amsmath}
\usepackage{amssymb}
\usepackage{booktabs}
\usepackage{multirow}
\usepackage{amssymb}
\usepackage{lineno,hyperref}
\usepackage{algorithm}
\usepackage{algpseudocode}
\usepackage{graphicx}
\usepackage{rotating}
\usepackage[justification=centering]{caption}
\usepackage{caption}
\usepackage{subcaption}
\usepackage{multirow}
\usepackage{etoolbox}
\usepackage{array}
\usepackage{subcaption}
\usepackage{ amssymb }
\usepackage{multirow}
\usepackage{comment}
\usepackage{subcaption}
\usepackage{placeins}
\usepackage{bm}
\usepackage{listings}
\usepackage{xcolor}
\usepackage[labelformat=simple]{subcaption}

\definecolor{codegreen}{rgb}{0,0,1}
\definecolor{codegray}{rgb}{0.95,0.5,0.5}
\definecolor{codepurple}{rgb}{0.58,0,0.82}
\definecolor{backcolour}{rgb}{1, 1, 1}

\def\wacvPaperID{} % *** Enter the WACV Paper ID here
\def\confName{WACV}
\def\confYear{2025}

\begin{document}

\title{BadScan: An Architectural Backdoor Attack on Visual State Space Models}

\author{Om Suhas Deshmukh\\
IIT Dharwad\\
{\tt\small 210010033@iitdh.ac.in}
\and
Sankalp Nagaonkar\\
IIT Dharwad\\
{\tt\small 210020031@iitdh.ac.in}
\and
Achyut Mani Tripathi\\
IIT Dharwad\\
{\tt\small t.achyut@iitdh.ac.in}
\and
Ashish Mishra\\
HPE lab, Bangalore\\
{\tt\small mishraashish632@gmail.com}
}
\maketitle
%%%%%%%%% ABSTRACT
\begin{abstract}
   The newly introduced Visual State Space Model (VMamba), which employs \textit{State Space Mechanisms} (SSM) to interpret images as sequences of patches, has shown exceptional performance compared to Vision Transformers (ViT) across various computer vision tasks. However, recent studies have highlighted that deep models are susceptible to adversarial attacks. One common approach is to embed a trigger in the training data to retrain the model, causing it to misclassify data samples into a target class, a phenomenon known as a backdoor attack. In this paper, we first evaluate the robustness of the VMamba model against existing backdoor attacks. Based on this evaluation, we introduce a novel architectural backdoor attack, termed BadScan, designed to deceive the VMamba model. This attack utilizes bit plane slicing to create visually imperceptible backdoored images. During testing, if a trigger is detected by performing XOR operations between the $k^{th}$ bit planes of the modified triggered patches, the traditional 2D selective scan (SS2D) mechanism in the visual state space (VSS) block of VMamba is replaced with our newly designed BadScan block, which incorporates four newly developed scanning patterns. We demonstrate that the BadScan backdoor attack represents a significant threat to visual state space models and remains effective even after complete retraining from scratch. Experimental results on two widely used image classification datasets, CIFAR-10, and ImageNet-1K, reveal that while visual state space models generally exhibit robustness against current backdoor attacks, the BadScan attack is particularly effective, achieving a higher Triggered Accuracy Ratio (TAR) in misleading the VMamba model and its variants.
\end{abstract}
\section{Introduction}
\par 
Convolutional neural networks (CNNs) such as ResNet \cite{he2016deep}, Inception \cite{szegedy2016rethinking}, and EfficientNet \cite{huang2022tuberculosis}, along with Transformer-based models like ViT \cite{dosovitskiy2020image}, Conformer \cite{peng2021conformer}, Resformer \cite{tian2023resformer}, and MLP-mixer \cite{tolstikhin2021mlp}, have been widely applied to a range of computer vision tasks. These include object detection, image classification, anomaly detection, segmentation, and multimodal learning tasks. These models have consistently delivered state-of-the-art performance across various domains. The recently proposed Mamba models \cite{waleffe2024empirical,gu2023mamba}, inspired by state-space models (SSMs) \cite{kalman1960new}, have outperformed existing deep learning models and established new benchmarks in various NLP tasks. This success has generated interest in the computer vision community to investigate SSMs for visual tasks \cite{zhu2024vision}. In \cite{zhu2024vision}, the first visual state-space model VMamba was introduced, featuring a 2D selective scan (SS2D)-based visual state space (VSS) block that analyzes image patches in four directions to improve feature extraction. The VMamba \cite{zhu2024vision} model and its variants \cite{chen2024mim}, \cite{ma2024u}, \cite{yue2024medmamba}, \cite{zhao2024rs} have achieved state-of-the-art performance in numerous computer vision tasks, emerging as strong competitors to transformer-based models like ViTs. Unlike transformer-based models, which suffer from quadratic complexity, the VMamba model offers linear computational efficiency.
\par 
Despite their exceptional performance across various domains, the deep learning models mentioned above are currently facing the challenge of \textbf{backdoor attack}, wherein attackers deliberately corrupt the models by training them on manipulated datasets. These datasets contain hidden triggers embedded in input samples from a source class, which cause the network to misclassify these input samples as a different, often target, class. The majority of existing backdoor attack techniques operate in two main ways: first, by retraining a neural network on datasets containing samples with either visible or imperceptible triggers \cite{gu2019badnets},\cite{nguyen2021wanet} \cite{liu2020reflection}; second, by directly or indirectly modifying the model’s weight values through adversarial sampling \cite{feng2022fiba,nguyen2021wanet,wang2022invisible,zhong2020backdoor,zhao2022natural}. The attack methods mentioned earlier have a notable drawback: their backdoor effects can be easily countered by re-training the deep learning models. Furthermore, a model's accuracy is not only affected by its weights but also by its architecture, including the configuration and connections among its layers. To address this problem, \cite{bober2023architectural} introduced an innovative architectural backdoor attack. This method deceives an AlexNet model by injecting high values into the feature map vector following the average pooling layer, triggered by the presence of a hidden trigger in an input image. However, the adversarial images generated by this architectural backdoor attack are visually detectable and cannot be directly used against the VMamba method, as it targets the average pooling layer, which is absent in the VMamba model. Given the impressive performance of the VMamba model compared to ViT models, it is essential to assess its robustness against current backdoor attacks to ensure effective real-time application deployment. \\
To the best of our knowledge, the VMamba model's resistance to both training-based and architectural backdoor attacks has not been extensively investigated. Consequently, we first evaluated and benchmarked the VMamba model's performance against several prominent backdoor attacks, including BadNets \cite{gu2019badnets}, WaNet \cite{nguyen2021wanet}, and R-Fool \cite{liu2020reflection}. Additionally, we introduced a novel architectural backdoor attack that successfully deceives the VMamba model with a higher success rate than existing backdoor attack methods. Our proposed new attack is visually imperceptible, weight-agnostic, and remains effective even after re-training, unlike attacks such as BadNets \cite{gu2019badnets}, WaNet \cite{nguyen2021wanet}, and R-Fool \cite{liu2020reflection}.  
Our proposed backdoor attack significantly enhances stealth compared to current attack methods. \\  Our proposed method provides several notable contributions: Firstly, we introduced a novel weight-agnostic and imperceptible architectural backdoor attack specifically designed for the visual state space model, with the backdoor seamlessly integrated into the model's architecture. Secondly, we developed four distinct scanning patterns, Viz. Random Efficient Scan (RES), Random Efficient Addition Scan (REAS), Random Efficient Multiply Scan (REMS), and Random Efficient Dropping Scan (REDS), collectively known as "BadScan." These patterns function within the newly designed BadScan block to effectively deceive the VMamba model. Thirdly, we evaluated the robustness of the VMamba model against four backdoor attacks: BadNets, WaNet, R-Fool, and our proposed architectural backdoor attack (BadScan). Additionally, we compared the VMamba model’s robustness with that of other deep models, including ViT, CNN-based models, and MLP-mixer, under the same attacks. Finally, we demonstrated that the BadScan-based attack not only survives retraining but also achieves a higher Triggered Accuracy Ratio (TRA) compared to three state-of-the-art backdoor attacks across multiple datasets.

\section{Related Work}
\subsection{SSM in Vision} The exceptional performance of state space models in handling long sequences has led researchers across various fields to leverage these models for establishing new benchmarks. Recent state space models such as Vision Mamba \cite{zhu2024vision}, U-Mamba\cite{ma2024u}, ViViM\cite{chen2024mim}, Medmamba\cite{yue2024medmamba}, and RS-Mamba\cite{zhao2024rs} have been introduced for applications in image classification, segmentation, video understanding, medical image diagnosis, and remote sensing. A comprehensive review of the latest state space models across different domains is available in \cite{xu2024survey}.
\subsection{Backdoor Attacks on VSS Models}
Chengbin et al. \cite{du2024understandingrobustnessvisualstate} conducted an empirical study on the robustness of these models against adversarial images, which are generated by introducing imperceptible noise using various techniques, such as FGSM \cite{du2024understandingrobustnessvisualstate}, BIM \cite{du2024understandingrobustnessvisualstate}, and C\&W \cite{du2024understandingrobustnessvisualstate}. Generating adversarial images usually necessitates access to model details such as logits and weights. In contrast, the effectiveness of visual state space models in defending against backdoor attacks has not been thoroughly investigated and requires comprehensive examination. \textbf{To address this gap, we assessed the VMamba model's \cite{zhu2024vision} vulnerability to several backdoor attacks \cite{gu2019badnets},\cite{nguyen2021wanet} \cite{liu2020reflection} and developed a novel architectural backdoor attack that is weight-agnostic, visually imperceptible, and remains effective even after retraining.} Additionally, we have developed four distinct scanning methods that successfully mislead the visual state space model (VMamba), resulting in a higher TRA compared to existing backdoor attacks when applied to VMamba. 
\subsection{Types of Backdoor Attacks}
Backdoor attack techniques can be divided into three categories: attacks with triggers, architectural attacks, and attacks without triggers. In attacks with triggers \cite{dumford2020backdooring,rakin2020tbt }, specific triggers are inserted into the network layers and are activated when the model encounters a particular input. In attacks without triggers, triggers are introduced into the training data as small noise patches or signals \cite{gu2019badnets},\cite{nguyen2021wanet}, \cite{liu2020reflection}. The model, trained on this poisoned dataset, misclassifies samples into a target class when it detects the trigger.  In \cite{bober2023architectural}, the authors proposed an architectural backdoor attack in which a trigger, such as a small white or checkerboard patch, is inserted into an image. When this trigger is detected, a predefined function significantly amplifies the value in the feature map after the average pooling layer of the AlexNet model. This method has certain drawbacks: the trigger is visually noticeable, and the attack cannot be executed if the model does not include an average pooling layer, which is the case for VMamba. \textbf{In contrast, the proposed attack method employs a visually imperceptible trigger and utilizes a bit-level trigger detection algorithm, making it faster compared to previous methods that rely on statistical techniques and are slower in trigger detection \cite{bober2023architectural}. Additionally, the proposed attack method is compatible with VMamba, overcoming the applicability limitation.}
\subsection{Defense Against Backdoor Attacks}
A common strategy to protect deep models from adversarial attacks is adversarial training \cite{goodfellow2014explaining}, which involves retraining the model on a dataset containing both clean and adversarial samples. However, specific methods to defend against backdoor attacks are still lacking. Pruning-based methods, trigger removal methods, and knowledge distillation-based methods are popular defense techniques for protecting deep models from backdoor attacks. Knowledge distillation (KD)-based techniques involve transferring knowledge from a teacher model trained on clean data to help the backdoored model forget information about the inserted triggers \cite{li2021neural,xia2022eliminating}. Trigger removal methods \cite{Wang2019NeuralCI,Zhou2024EvolutionaryTD} work at the input layer to eliminate triggers embedded in the input, transforming an attacked input into a clean one and preventing the model from displaying abnormal behavior. For example, \cite{guoscale} details a method that monitors consistency across the logits of the deep model to remove triggers from inputs. Pruning-based methods \cite{liu2018fine,Bajcsy2021BaselinePA} focus on eliminating inactive neurons during the classification of clean inputs, effectively erasing backdoor characteristics and reducing the size of the compromised model. In our work, we first subjected the VMamba model to both the proposed and existing backdoor attacks. Subsequently, we employed attention-blocking \cite{waseda2023closer}  and token-dropping \cite{waseda2023closer} defense techniques to protect the VMamba model from different backdoor attacks.

\section{Methodology}
\subsection{\textbf{State Space Model}}
The deep networks based on the  SSM \cite{kalman1960new} discussed earlier rely on a conventional continuous system that maps a 1-dimensional input function or sequence, denoted as $x(t) \in \mathbb{R}$, through intermediate latent states $u(t) \in \mathbb{R}$ to an output $y(t) \in \mathbb{R}$. This process is described by a linear Ordinary Differential Equation (ODE):

\begin{equation} u'(t) = \bm{E}u(t) + \bm{F} x(t) \end{equation} \begin{equation} y(t) = \bm{G}u(t) \end{equation}

Here, $\bm{E} \in \mathbb{R}^{M \times M}$ represents the state matrix, while $\bm{F} \in \mathbb{R}^{M \times 1}$ and $\bm{G} \in \mathbb{R}^{1 \times M}$ denote the projection parameters. To adapt this continuous system for deep learning scenarios, S4 and Mamba discretize it. This adaptation involves introducing a timescale parameter $\triangle$ and converting $\bm{E}$ and $\bm{F}$ into discrete parameters $\overline{\bm{E}}$ and $\overline{\bm{F}}$ using a fixed discretization method. Typically, the zero-order hold (ZOH) method is applied for discretization, defined as follows:

\begin{equation} \overline{\bm{E}} = \exp(\triangle \bm{E}) \end{equation} \begin{equation} \overline{\bm{F}} = (\triangle \bm{E})^{-1}(\exp(\triangle \bm{F}) - \bm{I}) \cdot \triangle \bm{F} \end{equation}

The discrete representation of this linear system is given by:

\begin{equation} u'(t) = \overline{\bm{E}}u(t) + \overline{\bm{F}}x(t) \end{equation} \begin{equation} y(t) = \bm{G}u(t) \end{equation} \begin{equation} \bm{v} = (\bm{G}\overline{\bm{F}}, \bm{G}\overline{\bm{E}}\overline{\bm{E}}, \ldots, \bm{G}\overline{\bm{E}}^{L-1}\overline{\bm{F}}) \end{equation} \begin{equation} y = c_1 \cdot \bm{v} \end{equation}

Here, $\bm{v} \in \mathbb{R}^{M}$ represents a structured convolutional kernel, and $M$ denotes the length of the input sequence $x$.
\subsection{BadScan Attack}The BadScan attack method designed to deceive the VMamba model operates in two phases: first, it involves generating and detecting a hidden trigger; second, it disrupts the scan pattern of a VSS block in the VMamba model by substituting the SS2D-based VSS block with the proposed BadScan block. The specifics of trigger detection and the BadScan block are outlined as follows: 
\subsubsection{Generation, Insertion and Detection of a Trigger}
\begin{figure}[htbp!]
        \centering
          \includegraphics[width=0.42\textwidth]{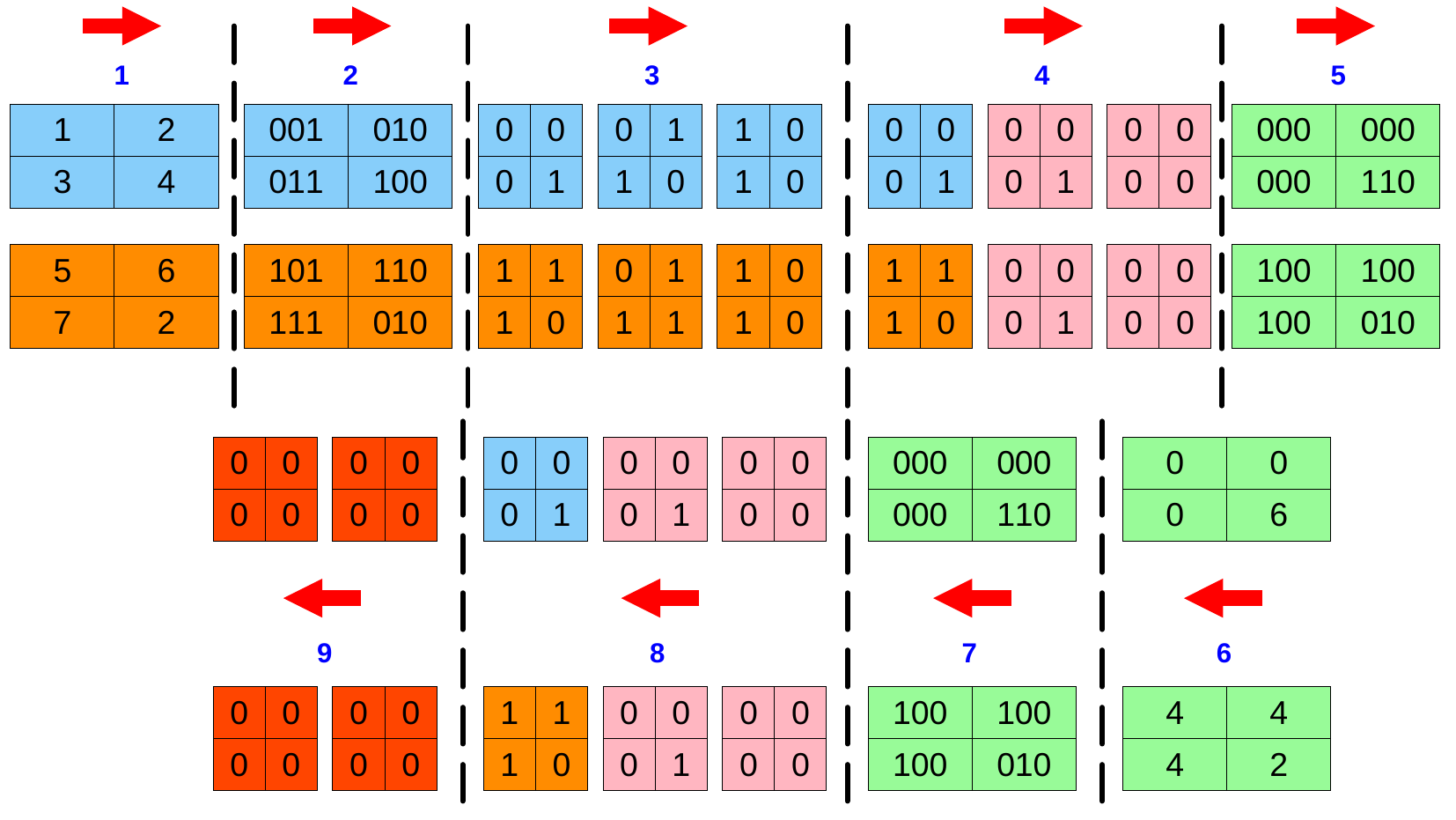}
        \caption{Workflow of Trigger Generation and Detection (Illustrated with a 3-Bit Representation for Better Clarity)}
        \label{XOR_Trigger} 
    \end{figure}

Figure \ref{XOR_Trigger} shows the process of creating and detecting triggered patches in an image using the bit plane slicing technique \cite{zhang2024learned}. 
In our work, the triggers can be inserted at any two predefined locations within an image, which are known only to the attacker. The steps involved in the creation and detection of a hidden trigger in an image are as follows:
\begin{enumerate}
    \item Let $\bm{P}_{i}$ and $\bm{P}_{j}$ denote the two patches chosen from two predefined locations in an image. The bit planes $\bm{B}$ for these patches, obtained using the bit plane slicing function ($f_{S}$), can be represented as follows \textbf{(Steps 1, 2 and 3 of Figure \ref{XOR_Trigger}):} 
    \begin{equation}
    \bm{B}_{7}^{i},\bm{B}_{6}^{i},...,\bm{B}_{1}^{i}, \bm{B}_{0}^{i}=f_{S}(\bm{P_{i}})
\end{equation}
\begin{equation}
    \bm{B}_{7}^{j},\bm{B}_{6}^{j},...,\bm{B}_{1}^{j}, \bm{B}_{0}^{j}=f_{S}(\bm{P_{j}})
\end{equation}
 Here, the leftmost bit planes ($\bm{B}_{7}^{j}$) are the most significant bit planes (MSBPs), while the rightmost bit planes ($\bm{B}_{1}^{j}$) are the least significant bit planes (LSBPs). In this work, each pixel value is represented using eight bits, resulting in a total of eight-bit planes for each selected patch.
\item In the next step, a bitwise XOR operation is performed between the $k^{th}$ bit planes of patches $\bm{P_{i}}$ and $\bm{P_{j}}$. This process is outlined as follows \textbf{(Step 4 of Figure \ref{XOR_Trigger}):} 
\begin{equation}
   \overline{ \bm{B}}^{i}_{k}=\overline{\bm{B}}^{j}_{k}=\bm{B}^{i}_{k}\oplus \bm{B}^{j}_{k}
\end{equation}

\item The next step involves generating the triggered patches $\bm{P}_{i}^{'}$ and $\bm{P}_{j}^{'}$ by replacing the $k^{th}$ bit plane $\bm{B}^{i}_{k}$ of patch $\bm{P}_{i}$ with its modified $k^{th}$ bit plane, $\overline{\bm{B}}^{i}_{k}$. A similar modification is applied to the $k^{th}$ bit plane $\bm{B}^{j}_{k}$ of patch $\bm{P}_{j}$. Consequently, the modified triggered patches $\bm{P}_{i}^{'}$ and $\bm{P}_{j}^{'}$ can be represented as follows \textbf{(Step 5 of Figure \ref{XOR_Trigger}):} 

\begin{equation}
     \bm{P}_{i}^{'}=f_{S}^{-1}(\bm{B}_{7}^{i},\bm{B}_{6}^{i},...,\overline{\bm{B}}_{1}^{i}, \overline{\bm{B}}_{0}^{i})
\end{equation}
\begin{equation}
     \bm{P}_{j}^{'}=f_{S}^{-1}(\bm{B}_{7}^{j},\bm{B}_{6}^{j},...,\overline{\bm{B}}_{1}^{j}, \overline{\bm{B}}_{0}^{j})
\end{equation}
Here $f_{S}^{-1}$ indicates the reverse operation of the bit plane slicing.
\item The modified patches ($\bm{P}_{i}^{'}$ and $\bm{P}_{j}^{'}$) are placed into the predefined locations of the original patches ($\bm{P}_{i}$ and $\bm{P}_{j}$) to generate a visually imperceptible attacked image from the clean image. 

\item To detect the trigger, the $k^{th}$ bit planes of the $i^{th}$ and $j^{th}$ patches from predefined locations in the input image are selected, and a bitwise XOR operation is performed between them to generate the $k^{th}$ trigger matrix $\bm{T}$. If the resulting trigger matrix is a zero matrix, the trigger is detected; otherwise, the trigger is absent from the input image. The trigger matrix $\bm{T}_{k}$ is computed using the following expression \textbf{ (Steps 6, 7, 8 and 9 of Figure \ref{XOR_Trigger}):} 
\begin{equation}
    \bm{T}_{k}=\overline{\bm{B}}_{k}^{i} \oplus \overline{\bm{B}}_{k}^{j}
\end{equation}
\end{enumerate}
 
\subsubsection{BadScan Block}
The main idea behind the proposed architectural backdoor attack is to interfere with the scanning pattern of the visual state space (VSS) block \cite{zhu2024vision} in the VMamba model upon detection of a trigger within the input image.
 \begin{figure}[htbp!]
        \centering
        \includegraphics[width=8cm, height=4.5cm]{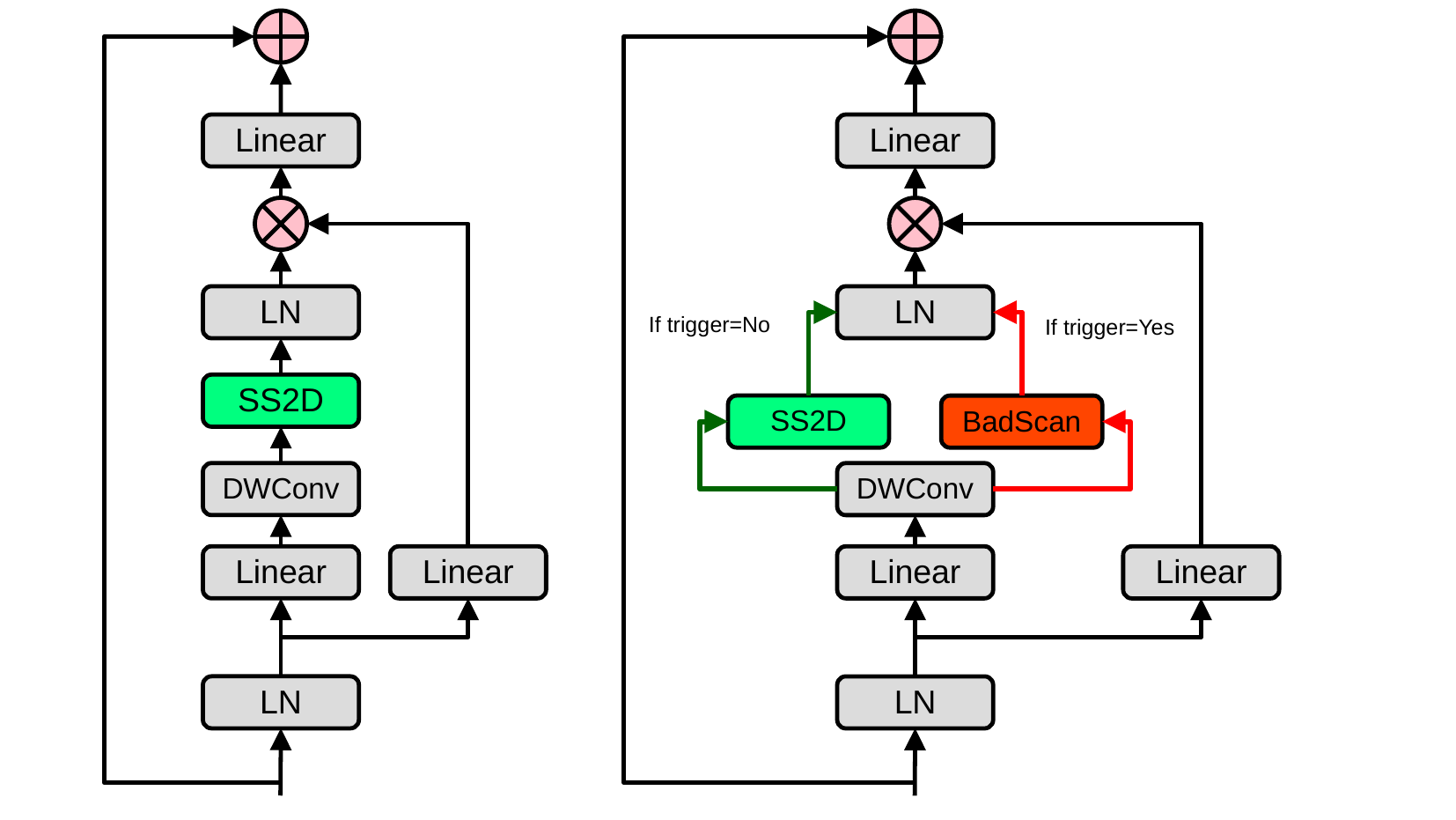}
        \caption{SS2D-Based VSS Block (Left) \cite{zhu2024vision} Vs BadScan Block (Right) }
        \label{BadScan_Block}
    \end{figure}
 Figure \ref{BadScan_Block} illustrates the difference between the standard VSS block and the VSS block affected by the proposed attack. In the absence of an imperceptible hidden trigger, patch scanning proceeds using the standard SS2D-based VSS block (indicated in green). However, when an imperceptible hidden trigger is present, the proposed BadScan block (indicated in red) takes over the scanning process. 
\begin{figure}[htbp!]
        \centering 
        \includegraphics[width=0.42\textwidth]{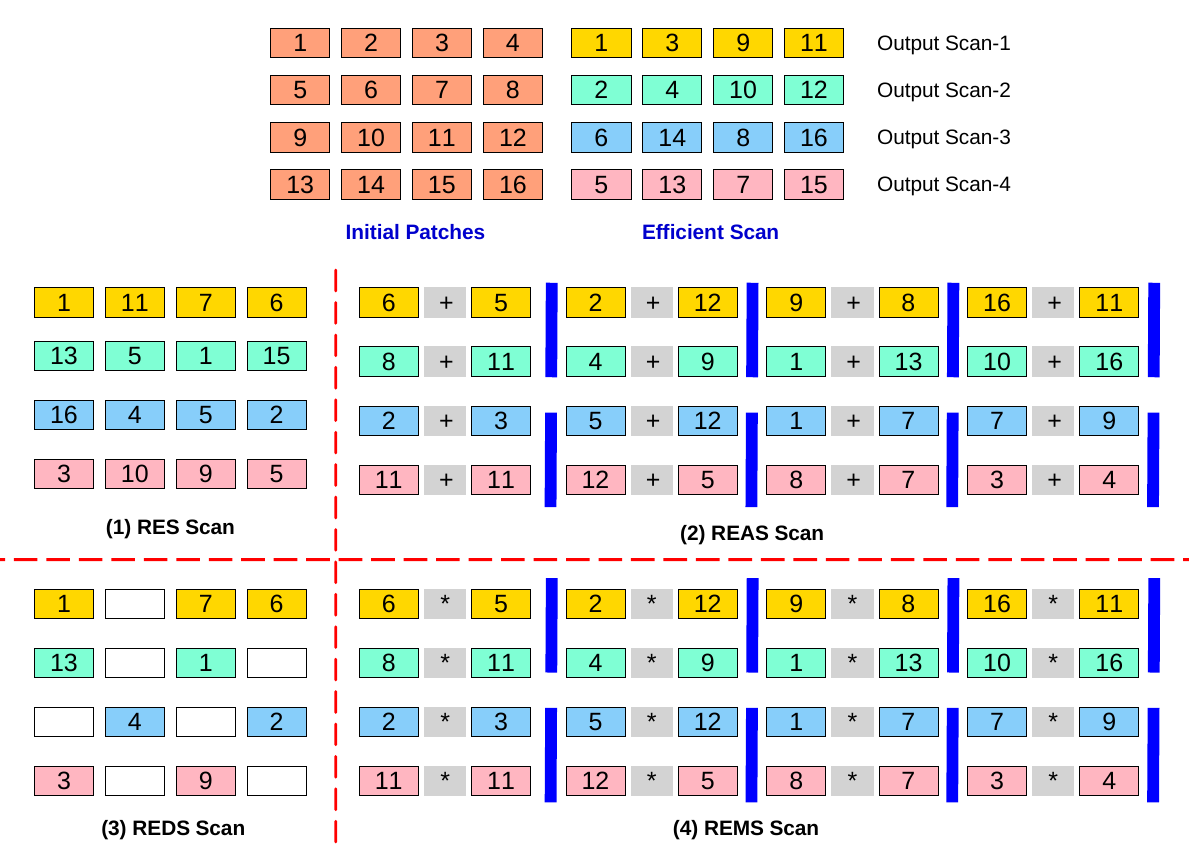}
        \caption{Working of RES, REAS, REMS and REDS Scanning Patterns Inside BadScan Block }
        \label{Four_Scan}
    \end{figure}
 Figure \ref{Four_Scan} illustrates the working of the four proposed scanning patterns within the BadScan block for a given set of image patches.     
The details of the four proposed scanning patterns are as follows:

\begin{enumerate}
 
    \item \textbf{Random Efficient Scan (RES)}: 
    This scanning method employs an efficient scan \cite{pei2024efficientvmamba} as its base technique. The key distinction is that, in this approach, the patches corresponding to the four output scan patterns of the efficient scan are selected randomly, with repetition of patches permitted.
    \item \textbf{Random Efficient Addition Scan (REAS)}:
    This scanning method is similar to the RES scan, with the main difference being that each patch corresponding to the four output scan patterns is the result of adding any two randomly selected patches.
      \item \textbf{Random Efficient Patch Dropping Scan (REDS)}: This scanning method is similar to the RES scan but with a key difference: initially, each of the patches corresponding to the four output scan patterns is selected randomly. Later, \textbf{$r\%$} of the patches in each of the four output scan patterns are dropped. In our work the value of $r$ is set to 20\%.
     \item \textbf{Random Efficient Multiply Scan (REMS)}:
     This scanning method is similar to the RES scan, with the key difference being that each patch corresponding to the four output scan patterns is generated by elementwise multiplication of any two randomly selected patches. 
\end{enumerate}

\section{Experiments \& Results}
This section will detail the experimental setup and discuss the results obtained.
\subsection{Image Datasets}
We selected two widely used image classification datasets, CIFAR-10 \cite{krizhevsky2014cifar} and ImageNet-1K \cite{imagenet15russakovsky}, to evaluate the performance of the VMamba model against both existing and proposed backdoor attacks. The poison ratio is set at 0.33 when assessing VMamba's robustness against BadNets \cite{gu2019badnets}, WaNet \cite{nguyen2021wanet}, R-Fool \cite{liu2020reflection} attacks. All images are resized and transformed to a dimension of $224 \times 224$. We used the same set of source-target pairs as those in \cite{gu2019badnets} to ensure a fair comparison. The performance results for the BadNet, R-Fool, and WaNet attacks are averaged across the three sets of source-target pairs for both CIFAR-10 \cite{krizhevsky2014cifar} and ImageNet-1K \cite{imagenet15russakovsky}. The source-target pairs selected for each dataset are as follows: For CIFAR-10, the pairs are \textbf{(Cat, Truck), (Car, Dog), and (Deer, Ship)}. In the case of the ImageNet-1K dataset, the chosen pairs are \textbf{(Seat Belt, Computer), (Shih-Tzu, Greyhound Racing), and (Bell Pepper, Chess Master)}.
\subsection{Deep Networks}
 In addition to evaluating the VMamba model's \cite{zhu2024vision} robustness against the four backdoor attacks, we also compared its robustness with that of ResNet-50 \cite{he2016deep}, ResNet-18 \cite{he2016deep}, ViT \cite{dosovitskiy2020image}, and MLP-mixer \cite{tolstikhin2021mlp} architectures.
 \subsection{Experimental Details}
In our work, we utilized a system with 128GB of RAM, running Ubuntu 22.04 LTS, and equipped with an NVIDIA GeForce RTX 4090 GPU. All models were trained using the SGD optimizer with a momentum of 0.90 and a learning rate 0.001. The scripts were written in PyTorch 2.2.0, and the attacked models were trained for ten epochs. For the BadNets attack, the patch size was set to 30, with other hyperparameters following those specified in \cite{gu2019badnets}. For the proposed BadScan attack, the patch size was configured to $(4 \times 4)$ across all three channels. The patches were placed in the top-left and bottom-left corners of the image. 
\subsection{Evaluation} The robustness of the VMamba and other deep models is assessed using the three metrics \cite{bober2023architectural}, viz. \textbf{Clean Task Accuracy (CTA)}, \textbf{Triggered Task Accuracy (TTA)} and \textbf{Triggered Accuracy Ratio (TAR)}. 
 The CTA is defined as the accuracy of a model when evaluated on clean test data samples. The TTA is defined as the accuracy of a model when evaluated on test data samples that contain triggers. 
The TAR is defined as the ratio of CTA to TTA. It indicates the relative decrease in performance that a backdoor attack causes when a deep model is fed input samples with triggers. A higher CTA combined with a lower TTA signifies an ideal backdoor attack that effectively deceives a deep model. Similarly, a high TAR value also indicates a highly effective backdoor attack.
 \subsection{VMamba Under Backdoor Attacks}
Table \ref{Performance_SOTA_VMamba_ImageNet} illustrates the performance of various deep models against different backdoor attacks.
 \begin{table}[htbp!]
\caption{Performance of Deep Models Against Backdoor Attacks}
\label{Performance_SOTA_VMamba_ImageNet}
\centering
\scalebox{0.50}{
\begin{tabular}{c|c|ccc|ccc}
\hline
\multirow{2}{*}{\textbf{Attacks}} & \multirow{2}{*}{\textbf{Model}} & \multicolumn{3}{c|}{\textbf{ImageNet-1K}}                                               & \multicolumn{3}{c}{\textbf{CIFAR-10}}                                            \\ \cline{3-8} 
                                  &                                 & \multicolumn{1}{c|}{\textbf{CTA}} & \multicolumn{1}{c|}{\textbf{TTA}} & \textbf{TAR} & \multicolumn{1}{c|}{\textbf{CTA}} & \multicolumn{1}{c|}{\textbf{TTA}} & \textbf{TAR} \\ \hline
\multirow{5}{*}{\begin{turn}{90}BadNets \cite{gu2019badnets}\end{turn}}                 & ResNet-18                       & \multicolumn{1}{c|}{72.03}             & \multicolumn{1}{c|}{46.27}             & 1.56              & \multicolumn{1}{c|}{85.43}             & \multicolumn{1}{c|}{46.77}             & 1.83             \\ \cline{2-8} 
                                  & ResNet-50                       & \multicolumn{1}{c|}{76.00}             & \multicolumn{1}{c|}{47.80}             & 1.59              & \multicolumn{1}{c|}{83.97}             & \multicolumn{1}{c|}{15.13}             & 5.55            \\ \cline{2-8} 
                                  & MLP-mixer                       & \multicolumn{1}{c|}{70.40}             & \multicolumn{1}{c|}{50.83}             &1.38              & \multicolumn{1}{c|}{95.17}             & \multicolumn{1}{c|}{7.00}             &  13.60           \\ \cline{2-8} 
                                  & ViT-S                       & \multicolumn{1}{c|}{70.37}             & \multicolumn{1}{c|}{46.67}             &1.51              & \multicolumn{1}{c|}{94.40}             & \multicolumn{1}{c|}{50.33}             &  1.88            \\ \cline{2-8} 
                                  & VMamba                   & \multicolumn{1}{c|}{66.13 \textcolor{blue}{}}            & \multicolumn{1}{c|}{{49.80}}             & \textbf{\textcolor{blue}{1.33}  $\uparrow$ (\textcolor{red}{14.17})}            & \multicolumn{1}{c|}{{93.40}}             & \multicolumn{1}{c|}{{66.53}}             & \textbf{\textcolor{blue}{1.40} $\uparrow$ (\textcolor{red}{4.87})}           \\ \hline
\multirow{5}{*}{\begin{turn}{90}WanNet \cite{nguyen2021wanet}\end{turn}}                 & ResNet-18                       & \multicolumn{1}{c|}{72.13}             & \multicolumn{1}{c|}{54.07}             & 1.33            & \multicolumn{1}{c|}{84.47}             & \multicolumn{1}{c|}{52.23}             & 1.61             \\ \cline{2-8} 
                                 & ResNet-50                       & \multicolumn{1}{c|}{75.20}             & \multicolumn{1}{c|}{53.53}             & 1.40              & \multicolumn{1}{c|}{84.57}             & \multicolumn{1}{c|}{17.53}             &  4.82           \\ \cline{2-8} 
                                  & MLP-mixer                       & \multicolumn{1}{c|}{66.67}             & \multicolumn{1}{c|}{38.93}             & 1.71             & \multicolumn{1}{c|}{91.93}             & \multicolumn{1}{c|}{71.37}             &      1.29       \\ \cline{2-8} 
                                  & ViT-S                       & \multicolumn{1}{c|}{73.17}             & \multicolumn{1}{c|}{60.20}             &  1.22            & \multicolumn{1}{c|}{94.43}             & \multicolumn{1}{c|}{74.00}             &  1.28            \\ \cline{2-8} 
                                  & VMamba                   & \multicolumn{1}{c|}{ {67.38}}            & \multicolumn{1}{c|}{{59.20}}             & \textbf{\textcolor{blue}{1.14}  $\uparrow$ (\textcolor{red}{14.39})}            & \multicolumn{1}{c|}{{93.47}}             & \multicolumn{1}{c|}{{86.97}}             & \textbf{\textcolor{blue}{1.07} $\uparrow$ (\textcolor{red}{5.20})}           \\ \hline
\multirow{5}{*}{\begin{turn}{90} Refool \cite{liu2020reflection} \end{turn}}                 & ResNet-18                       & \multicolumn{1}{c|}{72.87}             & \multicolumn{1}{c|}{41.37}             &1.76             & \multicolumn{1}{c|}{87.73}             & \multicolumn{1}{c|}{35.23}             &   2.49          \\ \cline{2-8} 
                                  & ResNet-50                       & \multicolumn{1}{c|}{75.40}             & \multicolumn{1}{c|}{48.97}             & 1.54              & \multicolumn{1}{c|}{83.73}             & \multicolumn{1}{c|}{8.37}             &  10.01           \\ \cline{2-8} 
                                  & MLP-mixer                       & \multicolumn{1}{c|}{69.10}             & \multicolumn{1}{c|}{38.90}             & 1.78             & \multicolumn{1}{c|}{94.30}             & \multicolumn{1}{c|}{32.57}             & 2.90            \\ \cline{2-8} 
                                  & ViT-S                       & \multicolumn{1}{c|}{72.63}             & \multicolumn{1}{c|}{49.47}             &  1.47            & \multicolumn{1}{c|}{96.47}             & \multicolumn{1}{c|}{38.20}             &      2.53        \\ \cline{2-8} 
                                  & VMamba                   & \multicolumn{1}{c|}{ {65.73}}            & \multicolumn{1}{c|}{{54.80}}             & \textbf{\textcolor{blue}{1.20}  $\uparrow$ (\textcolor{red}{14.30})}            & \multicolumn{1}{c|}{{94.93}}             & \multicolumn{1}{c|}{{50.73}}             & \textbf{\textcolor{blue}{1.87} $\uparrow$ (\textcolor{red}{4.40})}           \\ \hline
\multirow{3}{*}{\begin{turn}{90} \shortstack{BadScan} \end{turn}}

& VMamba                     & \multicolumn{1}{c|}{93.00}             & \multicolumn{1}{c|}{6.00}             &  \textbf{15.50}            & \multicolumn{1}{c|}{94.00}             & \multicolumn{1}{c|}{15.00}             &   \textbf{6.27}           \\ \cline{2-8} 

&MiM                      & \multicolumn{1}{c|}{90.00}             & \multicolumn{1}{c|}{6.00}             & \textbf{15.00}              & \multicolumn{1}{c|}{94.00}             & \multicolumn{1}{c|}{8.00}             &  \textbf{11.75}            \\ \cline{2-8}          

& EF-Mamba                    & \multicolumn{1}{c|}{90.00}             & \multicolumn{1}{c|}{5.00}             &  \textbf{18.00}           & \multicolumn{1}{c|}{98.00}             & \multicolumn{1}{c|}{11.00}             & \textbf{8.91}             \\ \cline{2-8}      
           
  \hline                                     
\end{tabular}
}
\end{table}
 For the VMamba model, the TAR values for BadNets, WaNet, R-Fool, and BadScan are 1.33, 1.14, 1.20, and 15.50, respectively, for the ImageNet-1K dataset. For the same attacks, the TAR values for the CIFAR-10 dataset are 1.40, 1.07, 1.87, and 6.27, respectively. It is evident that the VMamba model consistently achieves the lowest TAR values for BadNets, WaNet, and R-Fool attacks, indicating that it is more robust against these backdoor attacks compared to the other five deep models. Among the patch-based models, Viz. MLP-mixer, ViT, and V-Mamba, the robustness, measured by the TAR, is ranked in decreasing order as follows: V-Mamba, ViT, and MLP-mixer. This order remains consistent for BadNets, WaNet, and R-Fool attacks across both datasets. We also assessed the BadScan method's performance relative to other variants of the VMamba model, including Efficient-VMamba (EF-Mamba) \cite{pei2024efficientvmamba} and Mamba-in-Mamba (MiM) \cite{chen2024mim}. The BadScan method effectively deceives both the EF-Mamba and MiM models across the datasets, achieving a TAR comparable to that of the VMamba model.
\subsection{Ablation Study}
\subsubsection{Effectiveness in Other Domains}
In addition to assessing the performance of the proposed BadScan method in computer vision tasks, we also evaluated its effectiveness in other domains, such as audio classification. For this purpose, we used the EPIC-Sound Dataset \cite{EPICSOUNDS2023} and the Audio Mamba \cite{erol2024audio} model. The EPIC-Sound dataset includes audio recordings of various kitchen activities, such as washing and opening the fridge. For the EPIC-Sound dataset \cite{EPICSOUNDS2023}, the selected pairs include \textbf{(Cut Chop, Rustle), (Metal Collision, Drawer Open), and (Scrub, Tap Water)}. Table \ref{Impact_Bitplanes_AuM} displays the performance of the Audio Mamba model when subjected to the BadScan attack. The REDS scan-based BadScan method achieved the best results. This demonstrates that BadScan effectively deceives the Audio Mamba model, highlighting its potential applicability across different domains.

\begin{table}[htbp!]
\centering
\caption{BadScan Attack on Audio-Mamba For Different $(k)$}
\label{Impact_Bitplanes_AuM}
\scalebox{0.47}{
\begin{tabular}{c|c|cccc|cccc|cccc}
\hline
\multirow{2}{*}{\textbf{Dataset}}     & \multirow{2}{*}{\textbf{\begin{tabular}[c]{@{}c@{}}BadScan\\ Type\end{tabular}}} & \multicolumn{4}{c|}{\textbf{CTA}}                                                                                        & \multicolumn{4}{c|}{\textbf{TTA}}                                                                                        & \multicolumn{4}{c}{\textbf{TAR}}                                                                                        \\ \cline{3-14} 
                                      &                                                                                  & \multicolumn{1}{c|}{\textbf{K=1}} & \multicolumn{1}{c|}{\textbf{K=3}} & \multicolumn{1}{c|}{\textbf{K=5}} & \textbf{K=7} & \multicolumn{1}{c|}{\textbf{K=1}} & \multicolumn{1}{c|}{\textbf{K=3}} & \multicolumn{1}{c|}{\textbf{K=5}} & \textbf{K=7} & \multicolumn{1}{c|}{\textbf{K=1}} & \multicolumn{1}{c|}{\textbf{K=3}} & \multicolumn{1}{c|}{\textbf{K=5}} & \textbf{K=7} \\ \hline
\multirow{4}{*}{\begin{turn}{90}\textbf{EPIC-Sound}\end{turn}} & \textbf{RES}                                                                     & \multicolumn{1}{c|}{82.00}             & \multicolumn{1}{c|}{86.00}             & \multicolumn{1}{c|}{81.00}             &83.00              & \multicolumn{1}{c|}{19.00}             & \multicolumn{1}{c|}{20.00}             & \multicolumn{1}{c|}{15.00}             &  11.00           & \multicolumn{1}{c|}{4.32}             & \multicolumn{1}{c|}{4.30}             & \multicolumn{1}{c|}{5.40}             &  7.55            \\ \cline{2-14} 
                                      & \textbf{REAS}                                                                    & \multicolumn{1}{c|}{83.00}             & \multicolumn{1}{c|}{78.00}             & \multicolumn{1}{c|}{84.00}             & 85.00             & \multicolumn{1}{c|}{20.00}             & \multicolumn{1}{c|}{18.00}             & \multicolumn{1}{c|}{12.00}             & 20.00             & \multicolumn{1}{c|}{4.15}             & \multicolumn{1}{c|}{4.33}             & \multicolumn{1}{c|}{7.00}             &  4.25            \\ \cline{2-14} 
                                      & \textbf{REMS}                                                                    & \multicolumn{1}{c|}{82.00}             & \multicolumn{1}{c|}{75.00}             & \multicolumn{1}{c|}{84.00}             & 84.00             & \multicolumn{1}{c|}{15.00}             & \multicolumn{1}{c|}{20.00}             & \multicolumn{1}{c|}{20.00}             & 23.00             & \multicolumn{1}{c|}{5.47}             & \multicolumn{1}{c|}{3.75}             & \multicolumn{1}{c|}{4.20}             & 3.65             \\ \cline{2-14} 
                                      & \textbf{REDS}                                                                    & \multicolumn{1}{c|}{78.00}             & \multicolumn{1}{c|}{81.00}             & \multicolumn{1}{c|}{79.00}             & 84.00             & \multicolumn{1}{c|}{10.00}             & \multicolumn{1}{c|}{15.00}             & \multicolumn{1}{c|}{12.00}             & 20.00             & \multicolumn{1}{c|}{7.80}             & \multicolumn{1}{c|}{5.40}             & \multicolumn{1}{c|}{6.58}             & 4.20           \\ \hline
     
\end{tabular}}
\end{table}
\subsubsection{Impact of Number of Selected Bit Planes}
Tables \ref{Impact_Bitplanes}, \ref{Impact_Bitplanes_MiM}, and \ref{Impact_Bitplanes_EF} examine how the number of bit planes selected during trigger insertion affects the performance of the BadScan attack on the VMamba, MiM, and EF-Mamba models, respectively. 
\begin{table}[htbp!]
\centering
\caption{BadScan Attack on VMamba For Different $(k)$}
\label{Impact_Bitplanes}
\scalebox{0.47}{
\begin{tabular}{c|c|cccc|cccc|cccc}
\hline
\multirow{2}{*}{\textbf{Dataset}}     & \multirow{2}{*}{\textbf{\begin{tabular}[c]{@{}c@{}}BadScan\\ Type\end{tabular}}} & \multicolumn{4}{c|}{\textbf{CTA}}                                                                                        & \multicolumn{4}{c|}{\textbf{TTA}}                                                                                        & \multicolumn{4}{c}{\textbf{TAR}}                                                                                        \\ \cline{3-14} 
                                      &                                                                                  & \multicolumn{1}{c|}{\textbf{K=1}} & \multicolumn{1}{c|}{\textbf{K=3}} & \multicolumn{1}{c|}{\textbf{K=5}} & \textbf{K=7} & \multicolumn{1}{c|}{\textbf{K=1}} & \multicolumn{1}{c|}{\textbf{K=3}} & \multicolumn{1}{c|}{\textbf{K=5}} & \textbf{K=7} & \multicolumn{1}{c|}{\textbf{K=1}} & \multicolumn{1}{c|}{\textbf{K=3}} & \multicolumn{1}{c|}{\textbf{K=5}} & \textbf{K=7} \\ \hline
\multirow{4}{*}{\begin{turn}{90}\textbf{ImageNet-1K}\end{turn}} & \textbf{RES}                                                                     & \multicolumn{1}{c|}{96.00}             & \multicolumn{1}{c|}{89.00}             & \multicolumn{1}{c|}{85.00}             & 88.00             & \multicolumn{1}{c|}{15.00}             & \multicolumn{1}{c|}{5.00}             & \multicolumn{1}{c|}{9.00}             &  6.00            & \multicolumn{1}{c|}{6.4}             & \multicolumn{1}{c|}{17.80}             & \multicolumn{1}{c|}{9.44}             &14.67              \\ \cline{2-14} 
                                      & \textbf{REAS}                                                                    & \multicolumn{1}{c|}{94.00}             & \multicolumn{1}{c|}{90.00}             & \multicolumn{1}{c|}{89.00}             & 88.00             & \multicolumn{1}{c|}{12.00}             & \multicolumn{1}{c|}{16.00}             & \multicolumn{1}{c|}{8.00}             & 10.00             & \multicolumn{1}{c|}{7.83}             & \multicolumn{1}{c|}{5.63}             & \multicolumn{1}{c|}{11.13}             &  14.67            \\ \cline{2-14} 
                                      & \textbf{REMS}                                                                    & \multicolumn{1}{c|}{92.00}             & \multicolumn{1}{c|}{91.00}             & \multicolumn{1}{c|}{93.00}             &  87.00            & \multicolumn{1}{c|}{15.00}             & \multicolumn{1}{c|}{7.00}             & \multicolumn{1}{c|}{12.00}             & 6.00             & \multicolumn{1}{c|}{6.13}             & \multicolumn{1}{c|}{13.00}             & \multicolumn{1}{c|}{7.75}             & 14.50             \\ \cline{2-14} 
                                      & \textbf{REDS}                                                                    & \multicolumn{1}{c|}{90.00}             & \multicolumn{1}{c|}{91.00}             & \multicolumn{1}{c|}{93.00}             & 91.00             & \multicolumn{1}{c|}{7.00}             & \multicolumn{1}{c|}{8.00}             & \multicolumn{1}{c|}{6.00}             & 6.00             & \multicolumn{1}{c|}{12.86}             & \multicolumn{1}{c|}{11.38}             & \multicolumn{1}{c|}{15.50}             &  15.17            \\ \hline
\multirow{4}{*}{\begin{turn}{90} \textbf{CIFAR-10}\end{turn}}    & \textbf{RES}                                                                     & \multicolumn{1}{c|}{81.00}             & \multicolumn{1}{c|}{92.00}             & \multicolumn{1}{c|}{88.00}             & 87.00             & \multicolumn{1}{c|}{29.00}             & \multicolumn{1}{c|}{25.00}             & \multicolumn{1}{c|}{26.00}             & 17.00             & \multicolumn{1}{c|}{2.79}             & \multicolumn{1}{c|}{3.68}             & \multicolumn{1}{c|}{3.38}             &  5.12            \\ \cline{2-14} 
                                      & \textbf{REAS}                                                                    & \multicolumn{1}{c|}{83.00}             & \multicolumn{1}{c|}{89.00}             & \multicolumn{1}{c|}{91.00}             & 92.00             & \multicolumn{1}{c|}{17.00}             & \multicolumn{1}{c|}{15.00}             & \multicolumn{1}{c|}{20.00}             & 13.00             & \multicolumn{1}{c|}{4.88}             & \multicolumn{1}{c|}{5.93}             & \multicolumn{1}{c|}{4.55}             & 7.08             \\ \cline{2-14} 
                                      & \textbf{REMS}                                                                    & \multicolumn{1}{c|}{77.00}             & \multicolumn{1}{c|}{85.00}             & \multicolumn{1}{c|}{85.00}             &  84.00            & \multicolumn{1}{c|}{19.00}             & \multicolumn{1}{c|}{16.00}             & \multicolumn{1}{c|}{24.00}             &18.00              & \multicolumn{1}{c|}{4.05}             & \multicolumn{1}{c|}{6.27}             & \multicolumn{1}{c|}{3.54}             & 4.67             \\ \cline{2-14} 
                                      & \textbf{REDS}                                                                    & \multicolumn{1}{c|}{89.00}             & \multicolumn{1}{c|}{94.00}             & \multicolumn{1}{c|}{91.00}             & 89.00             & \multicolumn{1}{c|}{15.00}             & \multicolumn{1}{c|}{15.00}             & \multicolumn{1}{c|}{15.00}             & 15.00             & \multicolumn{1}{c|}{5.93}             & \multicolumn{1}{c|}{6.27}             & \multicolumn{1}{c|}{6.07}             & 5.93             \\ \hline
\end{tabular}}

\centering
\caption{BadScan Attack on MiM For Different $(k)$}
\label{Impact_Bitplanes_MiM}
\scalebox{0.47}{
\begin{tabular}{c|c|cccc|cccc|cccc}
\hline
\multirow{2}{*}{\textbf{Dataset}}     & \multirow{2}{*}{\textbf{\begin{tabular}[c]{@{}c@{}}BadScan\\ Type\end{tabular}}} & \multicolumn{4}{c|}{\textbf{CTA}}                                                                                        & \multicolumn{4}{c|}{\textbf{TTA}}                                                                                        & \multicolumn{4}{c}{\textbf{TAR}}                                                                                        \\ \cline{3-14} 
                                      &                                                                                  & \multicolumn{1}{c|}{\textbf{K=1}} & \multicolumn{1}{c|}{\textbf{K=3}} & \multicolumn{1}{c|}{\textbf{K=5}} & \textbf{K=7} & \multicolumn{1}{c|}{\textbf{K=1}} & \multicolumn{1}{c|}{\textbf{K=3}} & \multicolumn{1}{c|}{\textbf{K=5}} & \textbf{K=7} & \multicolumn{1}{c|}{\textbf{K=1}} & \multicolumn{1}{c|}{\textbf{K=3}} & \multicolumn{1}{c|}{\textbf{K=5}} & \textbf{K=7} \\ \hline
\multirow{4}{*}{\begin{turn}{90}\textbf{ImageNet-1K}\end{turn}} & \textbf{RES}                                                                     & \multicolumn{1}{c|}{92.00}             & \multicolumn{1}{c|}{89.00}             & \multicolumn{1}{c|}{87.00}             &  87.00            & \multicolumn{1}{c|}{10.00}             & \multicolumn{1}{c|}{5.00}             & \multicolumn{1}{c|}{8.00}             & 8.00             & \multicolumn{1}{c|}{9.20}             & \multicolumn{1}{c|}{17.80}             & \multicolumn{1}{c|}{10.88}             &  10.88            \\ \cline{2-14} 
                                      & \textbf{REAS}                                                                    & \multicolumn{1}{c|}{92.00}             & \multicolumn{1}{c|}{90.00}             & \multicolumn{1}{c|}{90.00}             &  89.00            & \multicolumn{1}{c|}{12.00}             & \multicolumn{1}{c|}{12.00}             & \multicolumn{1}{c|}{12.00}             &  15.00            & \multicolumn{1}{c|}{7.67}             & \multicolumn{1}{c|}{7.50}             & \multicolumn{1}{c|}{7.50}             & 5.93             \\ \cline{2-14} 
                                      & \textbf{REMS}                                                                    & \multicolumn{1}{c|}{91.00}             & \multicolumn{1}{c|}{88.00}             & \multicolumn{1}{c|}{90.00}             & 90.00             & \multicolumn{1}{c|}{11.00}             & \multicolumn{1}{c|}{6.00}             & \multicolumn{1}{c|}{15.00}             & 8.00             & \multicolumn{1}{c|}{8.27}             & \multicolumn{1}{c|}{14.67}             & \multicolumn{1}{c|}{6.00}             &  11.25            \\ \cline{2-14} 
                                      & \textbf{REDS}                                                                    & \multicolumn{1}{c|}{90.00}             & \multicolumn{1}{c|}{90.00}             & \multicolumn{1}{c|}{87.00}             & 95.00             & \multicolumn{1}{c|}{6.00}             & \multicolumn{1}{c|}{7.00}             & \multicolumn{1}{c|}{8.00}             & 10.00             & \multicolumn{1}{c|}{15.00}             & \multicolumn{1}{c|}{12.86}             & \multicolumn{1}{c|}{10.88}             & 9.50            \\ \hline
\multirow{4}{*}{\begin{turn}{90} \textbf{CIFAR-10}\end{turn}}    & \textbf{RES}                                                                     & \multicolumn{1}{c|}{90.00}             & \multicolumn{1}{c|}{96.00}             & \multicolumn{1}{c|}{91.00}             &  93.00            & \multicolumn{1}{c|}{8.00}             & \multicolumn{1}{c|}{13.00}             & \multicolumn{1}{c|}{20.00}             & 12.00            & \multicolumn{1}{c|}{11.25}             & \multicolumn{1}{c|}{7.38}             & \multicolumn{1}{c|}{4.55}             &  7.75            \\ \cline{2-14} 
                                      & \textbf{REAS}                                                                    & \multicolumn{1}{c|}{94.00}             & \multicolumn{1}{c|}{95.00}             & \multicolumn{1}{c|}{93.00}             &89.00              & \multicolumn{1}{c|}{8.00}             & \multicolumn{1}{c|}{11.00}             & \multicolumn{1}{c|}{9.00}             &  11.00            & \multicolumn{1}{c|}{11.75}             & \multicolumn{1}{c|}{8.64}             & \multicolumn{1}{c|}{10.33}             &  8.09            \\ \cline{2-14} 
                                      & \textbf{REMS}                                                                    & \multicolumn{1}{c|}{89.00}             & \multicolumn{1}{c|}{90.00}             & \multicolumn{1}{c|}{90.00}             &  86.00           & \multicolumn{1}{c|}{15.00}             & \multicolumn{1}{c|}{22.00}             & \multicolumn{1}{c|}{20.00}             &  16.00            & \multicolumn{1}{c|}{5.93}             & \multicolumn{1}{c|}{4.09}             & \multicolumn{1}{c|}{4.50}             & 5.38            \\ \cline{2-14} 
                                      & \textbf{REDS}                                                                    & \multicolumn{1}{c|}{88.00}             & \multicolumn{1}{c|}{93.00}             & \multicolumn{1}{c|}{89.00}             & 88.00             & \multicolumn{1}{c|}{16.00}             & \multicolumn{1}{c|}{11.00}             & \multicolumn{1}{c|}{15.00}             &  14.00           & \multicolumn{1}{c|}{5.50}             & \multicolumn{1}{c|}{8.45}             & \multicolumn{1}{c|}{5.93}             & 6.29             \\ \hline
\end{tabular}}
\centering
\caption{BadScan Attack on EF-Mamba For Different $(k)$}
\label{Impact_Bitplanes_EF}
\scalebox{0.47}{
\begin{tabular}{c|c|cccc|cccc|cccc}
\hline
\multirow{2}{*}{\textbf{Dataset}}     & \multirow{2}{*}{\textbf{\begin{tabular}[c]{@{}c@{}}BadScan\\ Type\end{tabular}}} & \multicolumn{4}{c|}{\textbf{CTA}}                                                                                        & \multicolumn{4}{c|}{\textbf{TTA}}                                                                                        & \multicolumn{4}{c}{\textbf{TAR}}                                                                                        \\ \cline{3-14} 
                                      &                                                                                  & \multicolumn{1}{c|}{\textbf{K=1}} & \multicolumn{1}{c|}{\textbf{K=3}} & \multicolumn{1}{c|}{\textbf{K=5}} & \textbf{K=7} & \multicolumn{1}{c|}{\textbf{K=1}} & \multicolumn{1}{c|}{\textbf{K=3}} & \multicolumn{1}{c|}{\textbf{K=5}} & \textbf{K=7} & \multicolumn{1}{c|}{\textbf{K=1}} & \multicolumn{1}{c|}{\textbf{K=3}} & \multicolumn{1}{c|}{\textbf{K=5}} & \textbf{K=7} \\ \hline
\multirow{4}{*}{\begin{turn}{90}\textbf{ImageNet-1K}\end{turn}} & \textbf{RES}                                                                     & \multicolumn{1}{c|}{92.00}             & \multicolumn{1}{c|}{93.00}             & \multicolumn{1}{c|}{90.00}             & 94.00             & \multicolumn{1}{c|}{14.00}             & \multicolumn{1}{c|}{10.00}             & \multicolumn{1}{c|}{15.00}             &6.00              & \multicolumn{1}{c|}{6.57}             & \multicolumn{1}{c|}{9.30}             & \multicolumn{1}{c|}{6.00}             &  14.67            \\ \cline{2-14} 
                                      & \textbf{REAS}                                                                    & \multicolumn{1}{c|}{97.00}             & \multicolumn{1}{c|}{96.00}             & \multicolumn{1}{c|}{95.00}             & 89.00             & \multicolumn{1}{c|}{15.00}             & \multicolumn{1}{c|}{10.00}             & \multicolumn{1}{c|}{15.00}             &  13.00            & \multicolumn{1}{c|}{6.47}             & \multicolumn{1}{c|}{9.60}             & \multicolumn{1}{c|}{6.33}             &6.85              \\ \cline{2-14} 
                                      & \textbf{REMS}                                                                    & \multicolumn{1}{c|}{92.00}             & \multicolumn{1}{c|}{92.00}             & \multicolumn{1}{c|}{96.00}             &  90.00            & \multicolumn{1}{c|}{12.00}             & \multicolumn{1}{c|}{9.00}             & \multicolumn{1}{c|}{10.00}             & 6.00             & \multicolumn{1}{c|}{7.67}             & \multicolumn{1}{c|}{10.22}             & \multicolumn{1}{c|}{9.60}             & 15.00             \\ \cline{2-14} 
                                      & \textbf{REDS}                                                                    & \multicolumn{1}{c|}{90.00}             & \multicolumn{1}{c|}{96.00}             & \multicolumn{1}{c|}{92.00}             & 94.00             & \multicolumn{1}{c|}{5.00}             & \multicolumn{1}{c|}{7.00}             & \multicolumn{1}{c|}{6.00}             & 10.00             & \multicolumn{1}{c|}{18.00}             & \multicolumn{1}{c|}{13.71}             & \multicolumn{1}{c|}{15.33}             &  9.40           \\ \hline
\multirow{4}{*}{\begin{turn}{90} \textbf{CIFAR-10}\end{turn}}    & \textbf{RES}                                                                     & \multicolumn{1}{c|}{78.00}             & \multicolumn{1}{c|}{85.00}             & \multicolumn{1}{c|}{84.00}             &  88.00            & \multicolumn{1}{c|}{28.00}             & \multicolumn{1}{c|}{31.00}             & \multicolumn{1}{c|}{19.00}             &  23.00           & \multicolumn{1}{c|}{2.79}             & \multicolumn{1}{c|}{2.74}             & \multicolumn{1}{c|}{4.42}             &   3.83           \\ \cline{2-14} 
                                      & \textbf{REAS}                                                                    & \multicolumn{1}{c|}{83.00}             & \multicolumn{1}{c|}{90.00}             & \multicolumn{1}{c|}{90.00}             & 77.00             & \multicolumn{1}{c|}{19.00}             & \multicolumn{1}{c|}{26.00}             & \multicolumn{1}{c|}{13.00}             &  17.00            & \multicolumn{1}{c|}{4.37}             & \multicolumn{1}{c|}{3.46}             & \multicolumn{1}{c|}{6.92}             &   5.41           \\ \cline{2-14} 
                                      & \textbf{REMS}                                                                    & \multicolumn{1}{c|}{78.00}             & \multicolumn{1}{c|}{81.00}             & \multicolumn{1}{c|}{77.00}             & 79.00            & \multicolumn{1}{c|}{20.00}             & \multicolumn{1}{c|}{20.00}             & \multicolumn{1}{c|}{30.00}             &  15.00            & \multicolumn{1}{c|}{3.90}             & \multicolumn{1}{c|}{2.57}             & \multicolumn{1}{c|}{2.57}             & 5.27            \\ \cline{2-14} 
                                      & \textbf{REDS}                                                                    & \multicolumn{1}{c|}{94.00}             & \multicolumn{1}{c|}{98.00}             & \multicolumn{1}{c|}{89.00}             &  90.00            & \multicolumn{1}{c|}{15.00}             & \multicolumn{1}{c|}{11.00}             & \multicolumn{1}{c|}{20.00}             & 13.00            & \multicolumn{1}{c|}{6.27}             & \multicolumn{1}{c|}{8.91}             & \multicolumn{1}{c|}{4.45}             & 6.92             \\ \hline
\end{tabular}}
\end{table}
Based on the TAR values for different $k$ values across both datasets and the VMamba model and its variants, the following observations can be made. For the VMamba and EF-Mamba models, the REDS scan-based BadScan outperformed the other three scan-based BadScan attacks in deceiving the VMamba and EF-Mamba models across both datasets. In the case of the MiM model, the best performance of the BadScan attack was achieved with the REDS scan for ImageNet-1K and the REAS scan for CIFAR-10. It is worth noting that, as evident from the tables mentioned above, regardless of the number of bit planes or scan type, the BadScan attack effectively deceives the VMamba and its variants across both datasets, achieving higher TAR values compared to BadNets, WaNets, and R-Fool attacks.

\subsection{Performance of Defense Methods}
Table \ref{Defense_Performance} illustrates the effectiveness of attention-blocking  \cite{waseda2023closer}  and token-dropping defense   \cite{waseda2023closer} methods against three different backdoor attacks when applied to the ViT and VMamba models on the ImageNet-1K dataset. 
\begin{table}[htbp!]
\caption{Performance of Different Defense Methods}
\label{Defense_Performance}
\centering
\scalebox{0.53}{
\begin{tabular}{c|ccc|ccc}
\hline
\textbf{Models}             & \multicolumn{3}{c|}{\textbf{ViT}}                                                             & \multicolumn{3}{c}{\textbf{VMamba}}                                                        \\ \hline
\textbf{Attacks}            & \multicolumn{1}{c|}{\textbf{BadNets}} & \multicolumn{1}{c|}{\textbf{WaNet}} & \textbf{R-Fool} & \multicolumn{1}{c|}{\textbf{BadNets}} & \multicolumn{1}{c|}{\textbf{WaNet}} & \textbf{R-Fool} \\ \hline
\textbf{Defense Method}     & \multicolumn{1}{c|}{\textbf{TTA}}     & \multicolumn{1}{c|}{\textbf{TTA}}   & \textbf{TTA}    & \multicolumn{1}{c|}{\textbf{TTA}}     & \multicolumn{1}{c|}{\textbf{TTA}}   & \textbf{TTA}    \\ \hline
\textbf{Attention Blocking} & \multicolumn{1}{c|}{55.00}            & \multicolumn{1}{c|}{92.33}           & 65.67           & \multicolumn{1}{c|}{76.33}            & \multicolumn{1}{c|}{60.00}          & 87.04           \\ \hline
\textbf{Token Dropping}     & \multicolumn{1}{c|}{46.00}            & \multicolumn{1}{c|}{91.00}           & 66.00           & \multicolumn{1}{c|}{74.20}            & \multicolumn{1}{c|}{86.00}          & 58.70          \\ \hline
\textbf{No Defense}         & \multicolumn{1}{c|}{46.67}            & \multicolumn{1}{c|}{60.20}          & 49.47           & \multicolumn{1}{c|}{49.80}            & \multicolumn{1}{c|}{59.20}          & 54.80           \\ \hline
\end{tabular}}
\end{table}
It is evident from Table \ref{Defense_Performance} that the TTA values increased significantly for both models across all three attacks. However, despite their effectiveness, these methods rely on the weight information of the backdoored model to secure the deep model, making them inadequate for defending against the proposed BadScan attack, which operates at the architectural level during test time. To address this, a potential area for future research could be the development of a defense mechanism specifically aimed at countering the BadScan attack.
\subsection{Attack Crafting Time}
The attack crafting time required by BadNets, WaNet, and R-Fool are  $5.72 e^{-6}$, $0.5325$ and $0.0119$ seconds, respectively. For the BadScan attack, the times required for $k$ values of  $1, 3, 5$, and $7$ are $0.000427, 0.000439, 0.000438$, and $0.000449$ seconds, respectively. Among these methods, BadNets is the fastest, while WaNet is the slowest. Notably, the proposed BadScan attack is the second fastest after BadNets and achieves the highest TAR in deceiving the VMamba model. Additionally, the time required for BadScan increases with the number of bit planes $(k)$ used to craft the backdoored sample. The reason for proposing our own trigger detection algorithm is that our detection method requires only 0.00036 seconds (approximately three times faster than the detection method described in \cite{bober2023architectural}), whereas the trigger detector in \cite{bober2023architectural} takes 0.0122 seconds. Additionally, unlike our trigger, which is visually imperceptible, the trigger in \cite{bober2023architectural} is visually detectable. 
\subsection{PSNR Values}
Maintaining a high PSNR value is essential for preserving the stealth of an attack. For ImageNet-1K, the PSNR values \cite{hore2010image} for the BadNets, WANet, R-Fool, and BadScan (REDS, $k=1$) attacks are 40.79, 38.09, 27.94, and 43.10, respectively. For CIFAR-10, the PSNR values for these same attacks are 43.75, 45.77, 28.50, and 49.11, respectively. Among the various attack methods, BadScan consistently achieves the highest PSNR values for both datasets, indicating that the hidden trigger introduced by BadScan remains visually imperceptible in the backdoored images.

\subsection{Persistence Against Retraining}We also evaluated the performance of the VMamba model under two conditions. First, we analyzed the model, initially trained on ImageNet-1K, after it was attacked with BadNet (using ImageNet-1K images) and then fine-tuned on the CIFAR-10 dataset. Second, we assessed the performance of the same attacked VMamba model after it was retrained from scratch on the CIFAR-10 dataset. Table \ref{Retraining_performance} presents a comparison of four attacks based on their effectiveness following the retraining of the VMamba model. The CTA and TTA values for the VMamba model remained comparable in both scenarios, indicating that the backdoor effects of BadNets, WaNet, and R-Fool were fully mitigated during either retraining from scratch or fine-tuning, as the backdoor-related weights were entirely removed. In contrast, the BadScan attack remained effective, significantly reducing the TTA value and achieving the highest TAR values of 9.79 and 6.27 for the first and second settings, respectively. This indicates that the proposed backdoor attack continues to be effective even after the model has been retrained or fine-tuned.
\begin{table}[htbp!]
\caption{Impact of Retraining and Fine-Tuning}
\label{Retraining_performance}
\centering
\scalebox{0.53}{
\begin{tabular}{c|ccc|ccc}
\hline
\multirow{2}{*}{\textbf{Attack}}       & \multicolumn{3}{c|}{\textbf{Fine-Tunning}}                                                    & \multicolumn{3}{c}{\textbf{From Scratch}}                                                    \\ \cline{2-7} 
                                       & \multicolumn{1}{c|}{\textbf{CTA}} & \multicolumn{1}{c|}{\textbf{TTA}} & \textbf{TAR}          & \multicolumn{1}{c|}{\textbf{CTA}} & \multicolumn{1}{c|}{\textbf{TTA}} & \textbf{TAR}          \\ \hline
\textbf{BadNets}                       & \multicolumn{1}{c|}{80.10}             & \multicolumn{1}{c|}{78.40}             & 1.02 $\uparrow$ (\textcolor{red}{8.77})                     & \multicolumn{1}{c|}{93.80}             & \multicolumn{1}{c|}{93.60}             & 1.01 $\uparrow$ (\textcolor{red}{5.26})                     \\ \hline
\textbf{WaNet}                         & \multicolumn{1}{c|}{80.50}             & \multicolumn{1}{c|}{82.80}             & 0.98 $\uparrow$ (\textcolor{red}{8.81})                     & \multicolumn{1}{c|}{93.40}             & \multicolumn{1}{c|}{93.80}             &0.99 $\uparrow$ (\textcolor{red}{5.28})                      \\ \hline
\textbf{R-Fool}                        & \multicolumn{1}{c|}{81.20}             & \multicolumn{1}{c|}{67.80}             &1.19 $\uparrow$ (\textcolor{red}{8.60})                      & \multicolumn{1}{c|}{94.00}             & \multicolumn{1}{c|}{93.80}             & 0.99 $\uparrow$ (\textcolor{red}{5.28})                     \\ \hline
\multicolumn{1}{c|}{\textbf{BadScan}} & \multicolumn{1}{c|}{91.10}             & \multicolumn{1}{c|}{9.26}             & \multicolumn{1}{c|}{\textbf{ \textcolor{blue}{9.79}}} & \multicolumn{1}{c|}{94.00}             & \multicolumn{1}{c|}{15.00}             & \multicolumn{1}{c}{\textbf{\textcolor{blue}{6.27}}} \\ \hline
\end{tabular}
}
\end{table}
\subsection{Qualitative Analysis}
\begin{figure*}[htbp!]
	\centering
		\begin{subfigure}[h]{0.14\textwidth}
			\includegraphics[width=2.5cm, height=1.4cm]{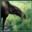}
			\caption{Clean}
			\label{A1}
		\end{subfigure}
  \begin{subfigure}[h]{0.14\textwidth}
			\includegraphics[width=2.5cm, height=1.4cm]{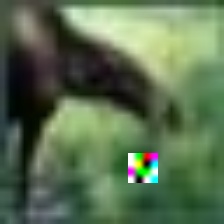}
			\caption{BadNets}
			\label{A2}
		\end{subfigure}
  \begin{subfigure}[h]{0.14\textwidth}
			\includegraphics[width=2.5cm, height=1.4cm]{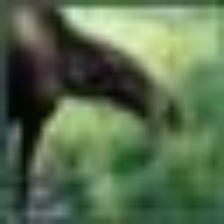}
			\caption{WaNet}
			\label{A3}
		\end{subfigure}
  \begin{subfigure}[h]{0.14\textwidth}
			\includegraphics[width=2.5cm, height=1.4cm]{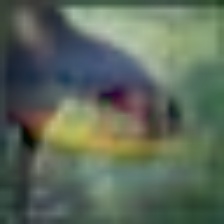}
			\caption{R-Fool}
			\label{A5}
		\end{subfigure}
  \begin{subfigure}[h]{0.14\textwidth}
			\includegraphics[width=2.5cm, height=1.4cm]{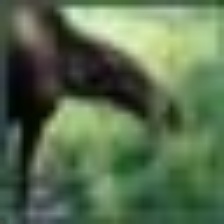}
			\caption{BadScan (REDS)}
			\label{A6}
		\end{subfigure}	
\caption{Clean and Attacked Images from CIFAR-10 (Target Class= Ships, Source Class= Deer)}\label{CIFAR-10_Attack_Images}
	\centering
		\begin{subfigure}[h]{0.14\textwidth}
			\includegraphics[width=2.5cm, height=1.4cm]{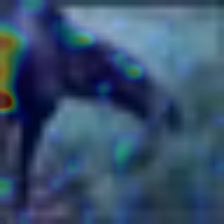}
			\caption{Clean}
			\label{A7}
		\end{subfigure}
  \begin{subfigure}[h]{0.14\textwidth}
			\includegraphics[width=2.5cm, height=1.4cm]{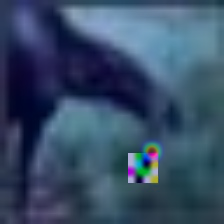}
			\caption{BadNets}
			\label{A8}
		\end{subfigure}
  \begin{subfigure}[h]{0.14\textwidth}
			\includegraphics[width=2.5cm, height=1.4cm]{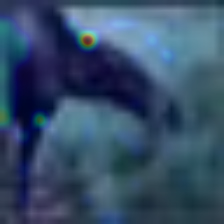}
			\caption{WaNet}
			\label{A9}
		\end{subfigure}
  \begin{subfigure}[h]{0.14\textwidth}
			\includegraphics[width=2.5cm, height=1.4cm]{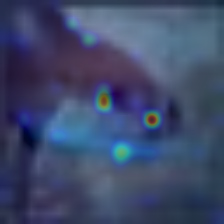}
			\caption{R-Fool}
			\label{A10}
		\end{subfigure}
  \begin{subfigure}[h]{0.14\textwidth}
			\includegraphics[width=2.5cm, height=1.4cm]{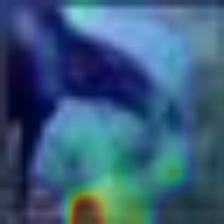}
			\caption{BadScan (REDS)}
			\label{A12}
		\end{subfigure}
\caption{Grad-CAM Maps of Clean and Attacked Images from CIFAR-10}\label{GCAM_CIFAR10}
	\centering
		\begin{subfigure}[h]{0.14\textwidth}
			\includegraphics[width=2.5cm, height=1.4cm]{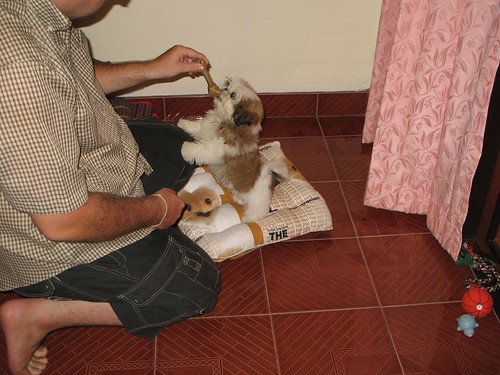}
			\caption{Clean}
			\label{A13}
		\end{subfigure}
  \begin{subfigure}[h]{0.14\textwidth}
			\includegraphics[width=2.5cm, height=1.4cm]{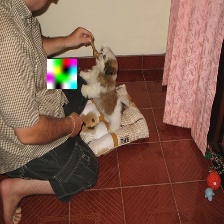}
			\caption{BadNets}
			\label{A14}
		\end{subfigure}
  \begin{subfigure}[h]{0.14\textwidth}
			\includegraphics[width=2.5cm, height=1.4cm]{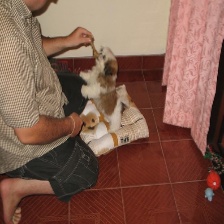}
			\caption{WaNet}
			\label{A15}
		\end{subfigure}
  \begin{subfigure}[h]{0.14\textwidth}
			\includegraphics[width=2.5cm, height=1.4cm]{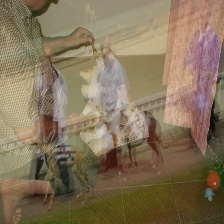}
			\caption{R-Fool}
			\label{A16}
		\end{subfigure}
  \begin{subfigure}[h]{0.14\textwidth}
			\includegraphics[width=2.5cm, height=1.4cm]{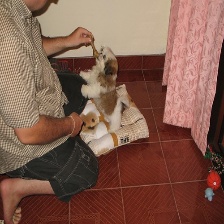}
			\caption{BadScan (REDS)}
			\label{A18}
		\end{subfigure}
\caption{ Clean and Attacked Images from ImageNet-1K (Target Class= Greyhound Racing, Source Class=Shih Tzu)}\label{ImageNet-1K Attacked Images}
	\centering
		\begin{subfigure}[h]{0.14\textwidth}
			\includegraphics[width=2.5cm, height=1.4cm]{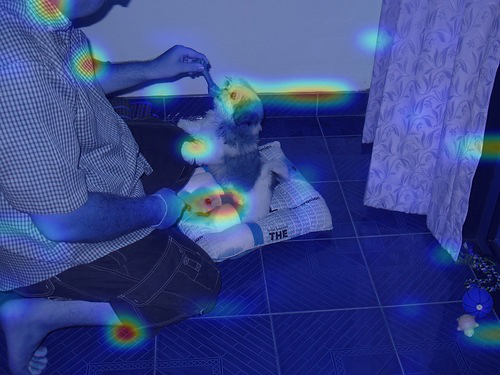}
			\caption{Clean}
			\label{A19}
		\end{subfigure}
  \begin{subfigure}[h]{0.14\textwidth}
			\includegraphics[width=2.5cm, height=1.4cm]{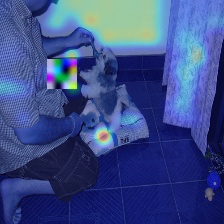}
			\caption{BadNets}
			\label{A20}
		\end{subfigure}
  \begin{subfigure}[h]{0.14\textwidth}
			\includegraphics[width=2.5cm, height=1.4cm]{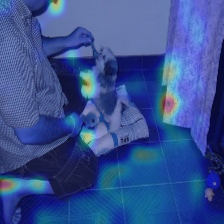}
			\caption{WaNet}
			\label{A21}
		\end{subfigure}
  \begin{subfigure}[h]{0.14\textwidth}
			\includegraphics[width=2.5cm, height=1.4cm]{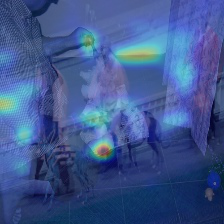}
			\caption{R-Fool}
			\label{A22}
		\end{subfigure}
  \begin{subfigure}[h]{0.14\textwidth}
			\includegraphics[width=2.5cm, height=1.4cm]{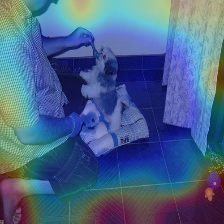}
			\caption{BadScan (REDS)}
			\label{A24}
		\end{subfigure}	
\caption{Grad-CAM Maps of Clean and Attacked Images from ImageNet-1K}\label{ImageNet-1K Attacked GCAM_Images}

\end{figure*}
The primary goal of a backdoor attack is to establish a strong association between an image and a concealed trigger, ensuring that the attacked model consistently classifies a source image into a target class whenever the hidden trigger is detected. For the CIFAR-10 and ImageNet-1K datasets, the source class is set to \textbf{Deer} and \textbf{Shih Tzu}, respectively. The target classes for these datasets are \textbf{Ships} and \textbf{Greyhound Racing}. Figures \ref{CIFAR-10_Attack_Images} and \ref{ImageNet-1K Attacked Images} illustrate the clean images and those with backdoor attacks for the CIFAR-10 and ImageNet-1K datasets, respectively. From these figures, it is evident that the hidden triggers inserted by Badnets, WaNets, and R-Fool are visually noticeable. In contrast, the hidden trigger crafted by BadScan is visually imperceptible across the backdoored images which is more preferable.
The effectiveness of an inserted trigger in fooling a model is demonstrated when the model starts focusing on the region where the trigger is present in a backdoored image. This behavior can be validated using Gradient-weighted Class Activation Mapping (Grad-CAM) plots, which illustrate the areas of focus for both clean and backdoored images. Figure \ref{GCAM_CIFAR10} and \ref{ImageNet-1K Attacked GCAM_Images} illustrate the Grad-CAM analysis of a clean and its backdoored image for the CIFAR-10 and its ImageNet-1K datasets, respectively for the VMamba network. Figure \ref{A7} illustrates that the model primarily focuses on the body region of a deer. In contrast, Figures \ref{A8}, \ref{A9}, and \ref{A10} show that the model focuses more on the regions with hidden triggers when subjected to the BadNets, WaNet, and R-Fool attacks. For the BadScan attack (Figure \ref{A12}), however, the model predominantly focuses on regions outside the deer's body, rather than on the regions where hidden triggers are present in the other attacks. Similar behavior is observed for the ImageNet-1K with the VMamba model across Figures \ref{A19}, \ref{A20}, \ref{A21} and \ref{A22}, corresponding to the scenarios with no attack, BadNet, WaNet, and R-Fool attacks, respectively. In the absence of an attack, the VMamba focuses more on the region with the Shih Tzu. However, with the BadNet, WaNet, and R-Fool attacks, the VMamba shifts its focus to regions containing the hidden trigger. For the BadScan attack (Figure \ref{A24}), the VMamba again highlights regions outside the Shih Tzu's body, rather than focusing on the regions where hidden triggers are present in the other attacks. 
In a successful backdoor attack, the model’s attention may shift to specific regions associated with the backdoor trigger, rather than focusing on the actual features of the source class. Grad-CAM visualizations can reveal this shift by showing that the model’s attention is redirected to features characteristic of the target class (e.g., focusing on areas resembling parts of a Ship instead of a Deer). These observations are reinforced by the Grad-CAM visualization of the VMamba model under the BadScan attack. Further analysis and results are provided in the supplementary material of this work.
\section{Conclusion}
In this paper, we introduce BadScan, a novel architectural backdoor attack aimed at deceiving the visual state space model. The proposed method utilizes bit plane slicing to embed a visually imperceptible hidden trigger within an image. A similar approach is used to detect the trigger within an input image, which then activates the BadScan attack if the trigger is found. The BadScan is weight-agnostic and retains its effectiveness even after the model undergoes retraining. Our experiments and results on three different datasets reveal two key findings: First, the VMamba model shows considerable robustness against existing backdoor attacks. Second, the proposed BadScan attack outperforms current backdoor attacks and effectively misleads the visual state space model with a high triggered accuracy ratio, thereby presenting a significant threat to the visual state space model and its variants. Furthermore, the BadScan attack effectively deceives the visual state space model even when applied to datasets from other domains, such as audio classification. The proposed BadScan attack has two limitations. First, it is currently untargeted, meaning it does not target specific classes or outcomes. Second, the attacker must be aware of the locations of the triggered patches within the images to execute the attack effectively. We hope that our proposed attack will inspire the vision community to create robust defense mechanisms, such as neural architecture search-based defenses or the designing of weight-agnostic networks, to secure visual state space models from BadScan and other advanced backdoor attack methods.
%%%%%%%%% REFERENCES
{\small
\bibliographystyle{ieee_fullname}
%\bibliography{output.bbl}
}

\end{document}